# A New Multifocus Image Fusion Method Using Contourlet Transform

**Fatemeh Vakili Moghadam. Hamid Reza Shahdoosti**

# روش جدید ادغام تصاویر چندزومه با استفاده از تبدیل کانتورلت

فاطمه وکیلی مقدم[1]، حمید رضا شاهدوستی

1- گروه مهندسی برق، دانشگاه صنعتی همدان

**Abstract**

A new multifocus image fusion approach is presented in this paper. First the contourlet transform is used to decompose the source images into different components. Then, some salient features are extracted from components. In order to extract salient features, spatial frequency is used. Subsequently, the best coefficients from the components are selected by the maximum selection rule. Finally, the inverse contourlet transform is applied to the selected coefficients. Experiments show the superiority of the proposed method.

**خلاصه**

در این مقاله تمرکز اصلی ما بر روی ادغام تصاویر چند زومه می‌باشد. ادغام تصاویر چندزومه به مفهوم ترکیب دو یا چند تصویر برای رسیدن به خروجی مطلوب می‌باشد. اخیراً روش‌های ادغام تصویر بسیاری برای تصاویر چندزومه به کار می‌رود. در این مقاله برای ادغام تصاویر از روش تبدیل موجک و تبدیل کانتورلت و چند روش ترکیبی استفاده شده است و در نهایت خروجی هرکدام از این الگوریتم ها مورد مقایسه قرار گرفته است.

**کلمات کلیدی**: ادغام تصاویر چند زومه، تبدیل موجک، تبدیل کانتورلت، تغییرات مکانی.

## 1 مقدمه

تصاویر، توصیف حقیقی از اشیا هستند. وقتی که تصاویر با دوربین گرفته می‌شوند محدودیت هایی از دوربین در آن‌ها دیده می‌شود. یکی از این محدودیت‌ها فاصله کانونی است که در این صورت تنها اشیایی که در فاصله کانونی دوربین می‌باشند به درستی دیده می‌شوند و مابقی تصاویر به صورت مات دیده می‌شوند ادغام تصاویر فرایند جمع آوری اطلاعات از چندین تصویر به یک تصویر ادغام شده برای فراهم آوردن توانایی های تفسیر بیشتر است. ادغام تصویر به طور وسیعی در پردازش تصاویر پزشکی، حوزه نظامی، سنجش از راه دور استفاده می‌شود. [1-3]

در سنجش از راه دور، ماهواره‌های کاوشگر زمینی اطلاعات تصویری را از قسمت‌های مختلف سطح زمین در زمان‌های مختلف تهیه می‌کنند برای بهره برداری موثر از این اطلاعات متنوع، از فناوری ادغام اطلاعات استفاده می‌شود. [4-3]

روش‌های ادغام اولیه، مستقیما، ادغام را بر روی تصاویر مرجع انجام می‌دهد، یکی از‌ساده‌ترین روش‌های ادغام متوسط پیکسل به پیکسل تصاویر مرجع را می‌گیرد. این روش ساده اغلب با کاهش کانتراست همراه است. با‌معرفی تبدیل هرم در سال 1975 روش‌های پیشرفته‌تری پدیدار شدند مشخص شد که اگر ادغام در حوزه تبدیل انجام شود، نتایج بهتری به دست می‌آید. با پیشرفت تئوری موجک، تجذیه‌ی موجک جای تجذیه هرمی را برای ادغام تصویر گرفت. [5] با پیشرفت تئوری کانتورلت نتایج بسیار عالی حاصل شد.

روش های مختلف و متنوعی که در ادغام تصاویر کاربرد دارند روش هایی مانند منطق فازی [6]، روش (LTM) که از یک مدل جریان با رویکرد چشمک‌زن روی دامنه شرلت (انسداد برشی)، ارائه می‌شود [7]،روش NSST[*] که بر مبنای تبدیل شرلت و پالس شبکه عصبی پیشنهاد شده است [8].
در این مقاله 4 روش برای ادغام تصاویر ارائه شده است.

## 2   ادغام تصاویر با استفاده از تبدیل موجک

### 2-1   تبدیل موجک دو بعدی گسسته

تبدیل موجک برای تجذیه سیگنال به مولفه‌های فرکانسی به کار می‌رود. در‌تبدیل موجک دوبعدی ابتدا دو فیلتر بالاگذر و پایین گذر را بر روی سطرها اعمال می‌کنیم و کاهش نرخ نمونه برداری انجام می‌دهیم. دو سیگنال با فرکانس‌های بالا و پایین ایجاد می‌شود و دوباره فیلتر بالاگذر و پایین‌گذر را بر روی ستون ها اعمال می‌کنیم و کاهش نرخ نمونه برداری انجام می‌دهیم در نهایت تصویر به 4 بخش فرکانسی تبدیل می‌شود.

### 2-2   گام های ادغام تصاویر با استفاده از تبدیل موجک [9]

برای ادغام تصاویر از دو تصویر ورودی A و B استفاده شده است.

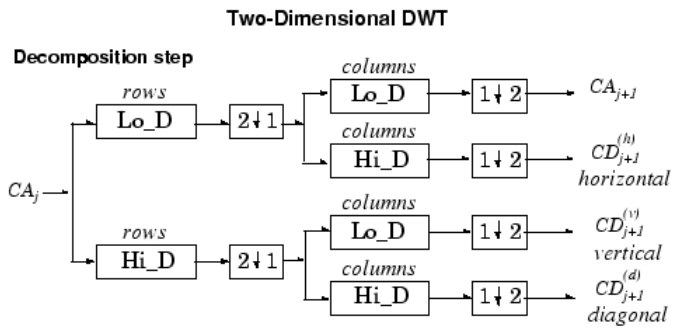
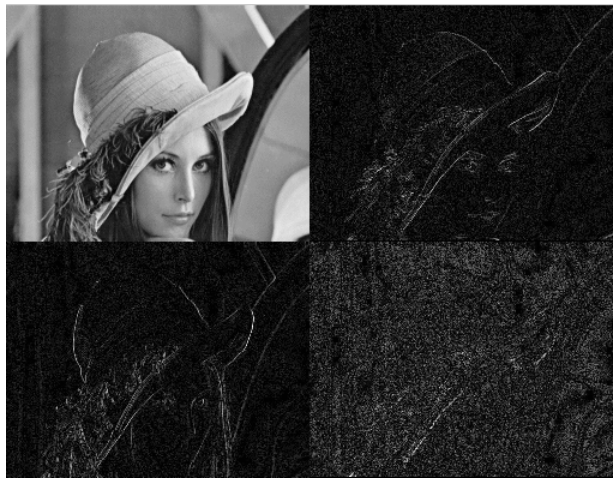

شکل 1- دیاگرام تبدیل موجک

شکل 2- نمایش ضرایب موجک

گام اول: محاسبه تبدیل موجک تصاویر ورودی

---

[*] -nonsubsampled shearlet transform

ضرایب موجک تصویر اول را با $[A_1, B_1, C_1, D_1]$ و تصویر دوم را با $[A_2, B_2, C_2, D_2]$ نمایش می‌دهیم.

گام دوم: ضرایب فرکانس بالای تصاویر ورودی را مقایسه می‌کنیم و مقدار ضریب بزرگتر را در تصویر خروجی قرار می‌دهیم. برای ضرایب فرکانس پایین، میانگین ضریب تصاویر ورودی گرفته می‌شود و در تصویر خروجی نمایش داده می‌شود.

$$\begin{cases} B_3 = B_1 & |B_1|>|B_2| \\ B_3 = B_2 & otherwise \end{cases} \begin{cases} C_3 = C_1 & |C_1|>|C_2| \\ C_3 = C_2 & otherwise \end{cases} \begin{cases} D_3 = D_1 & |D_1|>|D_2| \\ D_3 = D_2 & otherwise \end{cases}$$

$$A_3 = \frac{A_1 + A_2}{2}$$

(1)

گام سوم: در نهایت برای مشخص شدن تصویر ادغام شده عکس تبدیل موجک گرفته می‌شود.

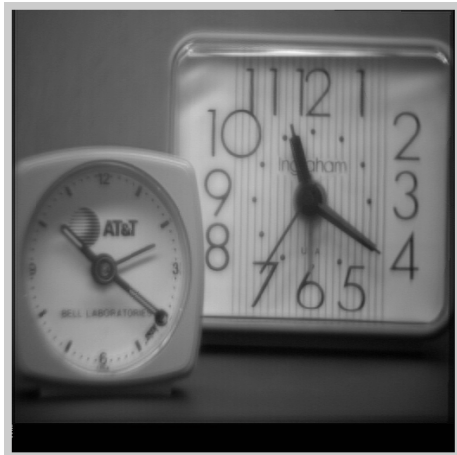

شکل ۳ - تصویر ادغام شده با روش تبدیل موجک

## 3  ادغام تصویر با استفاده از تغییرات مکانی[*] [10]

گام اول: تجذیه کردن تصاویر ورودی A و B به بلوک‌های M×N
بلوک‌های مربوط به تصویر A و B به ترتیب با $A_i$ و $B_i$ نمایش می‌دهیم.
M=N=8

گام دوم: محاسبه تغییرات مکانی(SF) برای هر بلوک

$$CF = \sqrt{\frac{1}{MN}\sum_{n=1}^{N}\sum_{m=1}^{M}(F(m,n)-F(m-1,n))^2} \quad RF = \sqrt{\frac{1}{MN}\sum_{n=1}^{N}\sum_{m=1}^{M}(F(m,n)-F(m,n-1))^2}$$

$$SF = \sqrt{RF^2 + CF^2}$$

(2)

تغییرات مکانی مربوط به $A_i$ و $B_i$ به ترتیب با $SF_i^A$ و $SF_i^B$ نمایش می‌دهیم.

---

[*] - spatial frequency

گام سوم: مقایسه تغییرات مکانی متناظر بلوک های $A_i$ و $B_i$

$$\begin{cases} A_i & SF_i^A > SF_i^B + TH \\ B_i & SF_i^A < SF_i^B - TH \\ (A_i + B_i)/2 & otherwise \end{cases} \quad (3)$$

TH یک آستانه است که برابر 1.75 تعریف می‌کنیم.

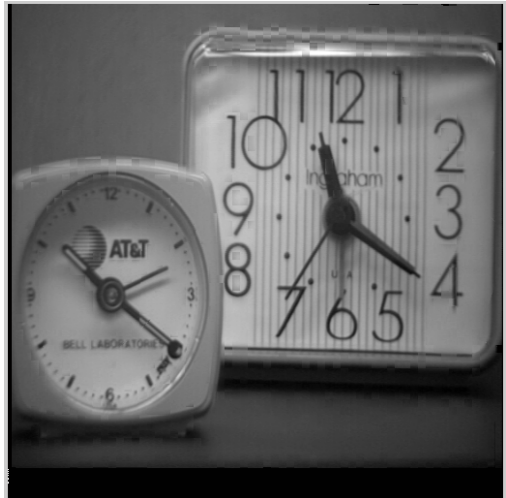

شکل 4 - تصویر ادغام شده با روش تغییرات مکانی

## 4  ادغام تصاویر با استفاده از تبدیل موجک و تغییرات مکانی

گام اول: محاسبه تبدیل موجک تصاویر ورودی [11]

گام دوم: تجزیه کردن ضرایب تبدیل موجک تصاویر ورودی A و B به بلوک‌های N×M

بلوک‌های مربوط به تصویر A شامل $D_1$ تا $D_4$.

بلوک‌های مربوط به تصویر B شامل $D_5$ تا $D_8$.

گام سوم: محاسبه تغییرات مکانی (SF) برای هر بلوک با استفاده از روابط CF و RF و SF.

گام چهارم: مقایسه تغییرات مکانی متناظر بلوک های ضرایب تبدیل موجک و از دو ضریب با فرکانس پایین میانگین گرفته می‌شود.

$$\begin{cases} F_i = D_i & SF_i^A > SF_i^B + TH \\ F_i = D_{i+4} & SF_i^A < SF_i^B - TH \\ F_i = (D_i + D_{i+4})/2 & otherwise \end{cases} \quad (4)$$

گام پنجم: در نهایت برای مشخص شدن تصویر ادغام شده عکس تبدیل موجک گرفته می‌شود.

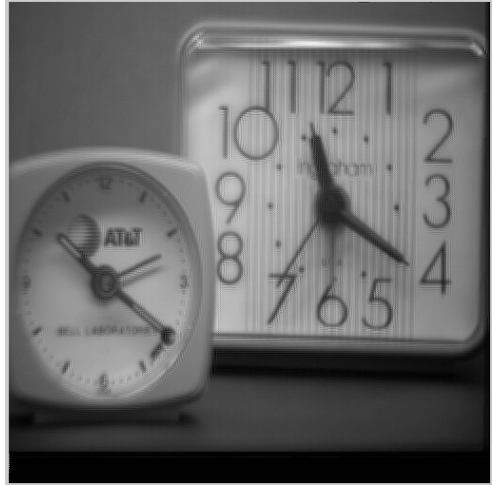

شکل 5- تصویر ادغام شده با روش تبدیل موجک و تغییرات مکانی

## 5   روش ادغام تصاویر با استفاده از تبدیل کانتورلت و تغییرات مکانی

### 5-1   تبدیل کانتورلت*

در تبدیل کانتورلت هم در فرکانس‌های مختلف و هم در جهت‌های مختلف تجذیه می‌شود و چون در جهت‌های مختلف تجذیه می‌شود نسبت به تبدیل موجک ضرایب تنک‌تری ایجاد می‌شود.

تبدیل کانتورلت در سال 2005 توسط دو و ویترلی پیشنهاد شد. در مرحله تجذیه از یک هرم لاپلاسین (LP) و یک فیلتر بانک جهتی (DFB) استفاده نموده. که درابتدا تصویر اصلی توسط هرم لاپلاسین به یک تصویر تقریب که تصویر پایین گذرنامیده شده و جزئیات تصویر که تصویر میان‌گذر نامیده شده تجذیه می‌شود و سپس توسط اعمال فیلتر بانک جهتی روی تصویر میان‌گذر اطلاعات جهتی به دست می‌آید. [12]

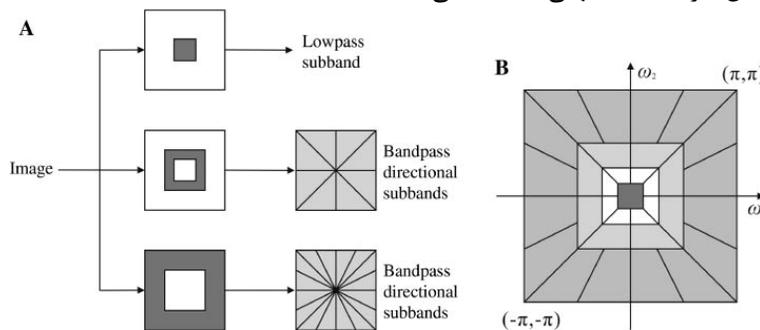

شکل 6 - دیاگرام تبدیل کانتورلت

### 5-2   گام های ادغام تصاویر با استفاده از تبدیل کانتورلت

گام اول: ابتدا تبدیل کانتورلت را برای تصاویر ورودی محاسبه می‌کنیم[13].

گام دوم: تجذیه کردن ضرایب تبدیل کانتورلت تصاویر ورودی A و B به بلوک‌های N×M بلوک‌های مربوط به تصویرA شامل $D_1$ تا $D_8$.

---

\* - contourlet transform

بلوک‌های مربوط به تصویر B شامل $D_9$ تا $D_{16}$.

گام سوم: محاسبه تغییرات مکانی(SF) برای هر بلوک با استفاده از روابط CF و RF و SF.

گام چهارم: مقایسه تغییرات مکانی متناظر بلوک های ضرایب تبدیل کانتورلت و از دو ضریب با فرکانس پایین میانگین گرفته می‌شود.

$$\begin{cases} F_i = D_i & SF_i^A > SF_i^B + TH \\ F_i = D_{i+8} & SF_i^A < SF_i^B - TH \\ F_i = (D_i + D_{i+8})/2 & otherwise \end{cases} \quad (5)$$

گام پنجم: در نهایت برای مشخص شدن تصویر ادغام شده عکس تبدیل کانتورلت گرفته می‌شود[14].

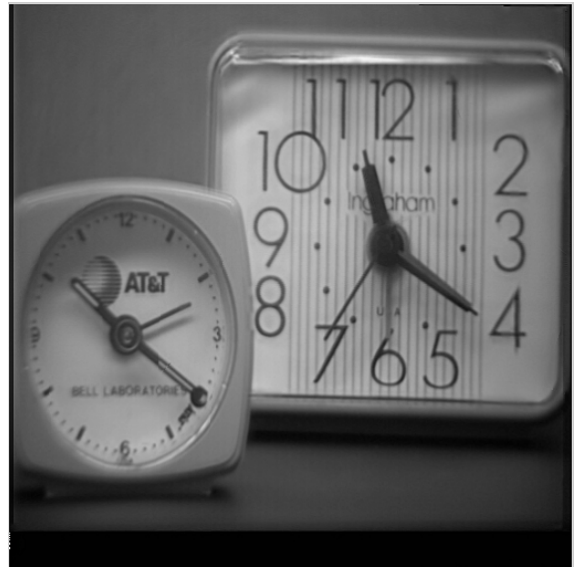

شکل 7 - تصویر ادغام شده با روش تبدیل کانتورلت

## 6 ارزیابی عددی

برای درک عددی اختلاف بین تصاویر ورودی و خروجی در روش های ادغام تصاویر مقدار خطای باید محاسبه گردد.

برای محاسبه این مقدار خطا از رابطه $RMSE^*$ استفاده می‌کنیم. به این مفهموم که ابتدا مقدار خطای تصویر اول با تصویر ادغام شده $RMSE_1$ و مقدار خطای تصویر دوم را با تصویر ادغام شده $RMSE_2$ به صورت مجزا محاسبه می‌کنیم و در نهایت میانگین گرفته می‌شود[15].

$$RMSE_1 = \sqrt{\frac{1}{I \times J} \sum_{i=1}^{I} \sum_{j=1}^{J} \left( A(i,j) - F(i,j) \right)^2}$$

(6)

$$RMSE_2 = \sqrt{\frac{1}{I \times J} \sum_{i=1}^{I} \sum_{j=1}^{J} \left( B(i,j) - F(i,j) \right)^2}$$

(7)

$$RMSE = \frac{RMSE_1 + RMSE_2}{2}$$

(8)

مقادیر $RMSE$ محاسبه شده برای چهار روش بالا در جدول زیر نمایش داده شده است.

همان طور که مشاهده می‌کنیم اگر چه روش تبدیل موجک مقدار خطای کمتری دارد و از نظر معیار عددی بهتر عمل کرده است و بعد روش تبدیل کانتورلت و بعد روش تغییرات مکانی و در نهایت روش تبدیل موجک و تغییرات مکانی مقدار خطای بالایی دارد، اما از نظر معیار چشمی روش تبدیل کانتورلت مناسب‌تر می‌باشد. این اختلاف نتایج به دلیل ضعفی است که در معیار ارزیابی عددی وجود دارد[16].

جدول ۱- ارزیابی عددی

| روش ادغام تصویر | RMSE |
|---|---|
| تبدیل موجک | 7.905 |
| تغییرات مکانی | 10.6426 |
| تبدیل موجک و تغییرات مکانی | 22.8594 |
| تبدیل کانتورلت و تغییرات مکانی | 9.0371 |

## 7 نتیجه گیری

در این مقاله یک روش جدید به منظور ادغام تصاویر چند زومه پیشنهاد شد. با استفاده از تغییرات فرکانسی در حوزه کانتورلت ضرایب برتر را استخراج کردیم و در نهایت با استفاده از قانون انتخاب ماکزیمم، ضرایب برتر را بدست آوردیم. روش پیشنهادی با چند روش متداول مانند روش موجک و روش پنجره مکانی مقایسه شدند و نشان دادیم که عملکرد روش پیشنهادی بهتر از سایر روش‌های موجود است.

---

[*] - Root Mean Square Error